\documentclass[runningheads]{llncs}

\usepackage{eccv}

\usepackage{eccvabbrv}
\usepackage{graphicx}
\usepackage{booktabs}
\usepackage{amsmath,amssymb}
\usepackage{multirow}
\usepackage{xcolor}
\usepackage{xspace}
\usepackage{iftex}
\ifpdftex
  \usepackage[accsupp]{axessibility}
\fi

\usepackage[breaklinks,colorlinks,citecolor=eccvblue]{hyperref}
\usepackage{orcidlink}

\newcommand{\real}{\textsc{real}\xspace}
\newcommand{\realsynth}{\textsc{realsynth}\xspace}
\newcommand{\kimodo}{\textsc{kimodo}\xspace}

\begin{document}

% ---------------------------------------------------------------
\title{Learning to Understand Body Language from Flight through Robust 3D Avatar Placing}

\titlerunning{Learning Body Language from Flight}

\author{Drago\c{s} Costea\inst{1} \and
Alina Marcu\inst{1} \and
Cristina Laz\u{a}r\inst{1} \and
Marius Leordeanu\inst{1,2,3}}
\authorrunning{D.~Costea et al.}
\institute{National University of Science and Technology ``Politehnica'' Bucharest, Romania\\
\email{\{alina.marcu,dragos.costea,cristina.lazar,marius.leordeanu\}@upb.ro} \and
``Simion Stoilow'' Institute of Mathematics of the Romanian Academy \and
NORCE Norwegian Research Centre AS}

\maketitle

\begin{abstract}
Perceiving human motion and intent at long range is a prerequisite for socially intelligent aerial robots, yet the data to learn it barely exists. We introduce \emph{Drones2BodyLanguage}, a dataset grounding human motion in real UAV footage: avatars manifesting ten communicative intents are placed into unmodified 4K drone scenes with metrically correct position, scale and orientation, maintained over hundreds of frames of camera motion. Enabling it is a lightweight geometric world model of the local scene -- semantically selected anchors lifted to 3D through streaming monocular depth -- in which a placement point is predicted as an affine anchor combination with provably rigid-invariant weights, and re-rendered under an SVD-fitted ground rotation. Across twelve architectures on scene- and motion-disjoint splits, training on placed data lifts mean intent accuracy by a wide margin for real, retargeted and generated motion alike, with gains confirmed on two in-the-wild scenes.
\end{abstract}

%-------------------------------------------------------------------------
\section{Introduction}
\label{sec:intro}

Non-verbal communication is the only channel left when a person signals to a rescue drone, or when a privacy-preserving camera must not record faces~\cite{perera2021uav,kim2025drone}. Communicative \emph{intent} can be recovered from 2D body keypoints alone in real time~\cite{costea2026nonverbalrealtimehumanaiinteraction}, but only at close range: the same study's aerial experiments show pose detection collapsing between $50$ and $100$\,m. Collecting labels at these distances is prohibitive, while fully synthetic rendering abandons the photometric statistics of real aerial video.

We take the middle road: keep the real scene, and \emph{place} the people. Compositing humans into unmodified drone footage preserves the true background, camera motion and pixel budget of a person at distance, with full control over label, distance and sample count. Doing this convincingly is a hard 3D tracking problem: a placed person must stand on walkable ground at the correct pixel height, stay attached to the same ground point while the drone pans, and rotate coherently with the scene -- yet that point lies on textureless asphalt. Our answer is a streaming geometric world model: semantically gated corners on world-static structure are lifted to 3D through noisy monocular depth, and the placement point is an affine combination of these anchors with weights \emph{exactly} invariant under rigid coordinate changes (Property~\ref{prop:invariance}) -- \emph{predicted}, not tracked, it survives texture loss, its depth sets the avatar's scale, and an SVD fit over the constellation supplies the per-frame ground rotation.

\smallskip
\noindent Our contributions are: \textbf{(1) Drones2BodyLanguage} -- $3{,}580$ detection-verified $80$-frame keypoint sequences over eight real drone scenes, three motion sources (real cut-outs \real, retargeted \realsynth, generated \kimodo~\cite{rempe2026kimodo}) and ten intents at $33$--$98$\,m (\cref{sec:dataset}); \textbf{(2) a novel 3D avatar placing algorithm} computing position, rotation and scale from noisy monocular depth and semantic segmentation (\cref{sec:method}); \textbf{(3) an intent-classification study at distance} over twelve architectures on scene- and motion-disjoint splits, with transfer confirmed on new real footage (\cref{sec:experiments}).

\begin{figure}[tb]
  \centering
  \includegraphics[width=0.98\linewidth]{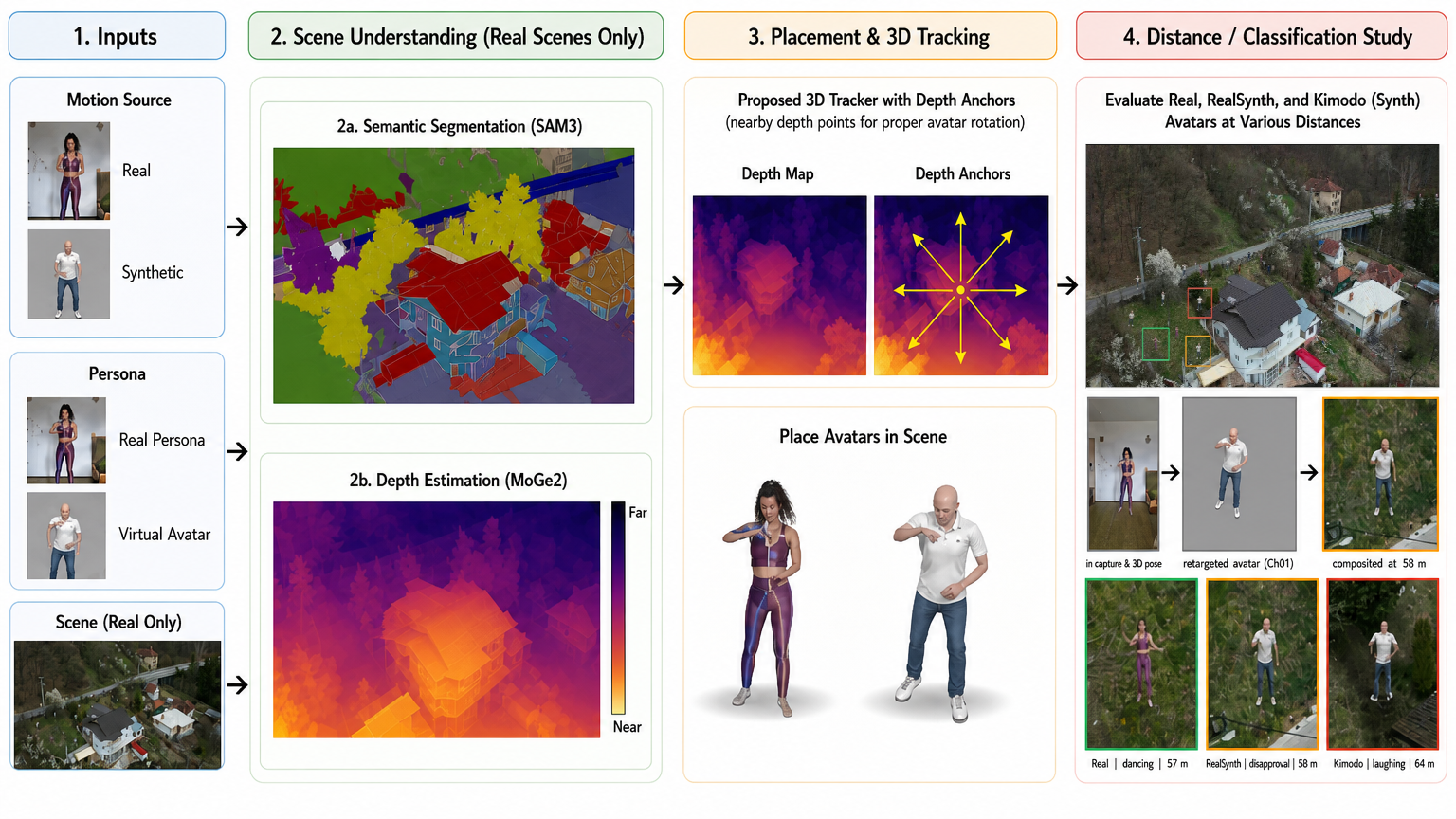}
  \caption{Overview: real and synthetic motions and personas are composited into real 4K drone scenes via SAM3 segmentation, metric depth and our 3D placement tracker; the \real{}/\realsynth{}/\kimodo{} placements drive the intent study at distance}
  \label{fig:teaser}
  \vspace{-5mm}

\end{figure}

%-------------------------------------------------------------------------
\section{Related Work}
\label{sec:related}

Image object/person placement -- adversarial transformers~\cite{lin2018stgan}, scene--object graphs~\cite{zhou2022graconet}, heatmap regression~\cite{zhu2023topnet}, plausibility and harmonization benchmarks~\cite{liu2021opa,cong2020dovenet}, cut-paste-learn~\cite{dwibedi2017cut,tripathi2019learning} and rendered people~\cite{varol2017learning} -- operates on single images or fully synthetic scenes; none addresses metric, rotation-consistent placement in unmodified aerial video. Point trackers, from pyramidal Lucas--Kanade~\cite{lucas1981iterative,shi1994good} to learned dense, long-range and 3D variants~\cite{doersch2023tapir,karaev2024cotracker,xiao2024spatialtracker,zhang2025tapip3d,xiao2025spatialtrackerv2}, and feedforward geometry engines~\cite{wang2025vggt,li2025megasam}, output point trajectories or global camera pose -- not the pose of a scene-anchored local frame around a deliberately \emph{textureless} target, which our anchor construction supplies in closed form, training-free~\cite{lepetit2009epnp}. Skeleton action recognition is mature~\cite{yan2018stgcn,chen2021ctrgcn,liu2020ntu120}, but communicative intent from 2D pose was benchmarked only recently and at close range~\cite{costea2026nonverbalrealtimehumanaiinteraction}. Our corpus isolates the synthetic-vs-real \emph{motion} gap~\cite{dai2024motionlcm,rempe2026kimodo} across three sources sharing scenes and compositing.

%-------------------------------------------------------------------------
\section{Method}
\label{sec:method}

Per scene (\cref{fig:teaser}): (1) scene preparation; (2) 3D-anchored ground-point tracking, outputting per placement and frame its 2D position, metric depth and ground rotation; (3) avatar rendering and compositing; (4) detection-verified harvesting. From each source video we extract a two-minute window maximizing near-camera walkable ground; every frame receives a streaming monocular depth map~\cite{lin2025depthanything3}, and SAM3~\cite{carion2025sam3} segments the first frame of each tracking epoch into a \emph{placeable} mask (road $\cup$ low vegetation, eroded) and an \emph{anchorable} mask that adds trees and buildings. Intrinsics use the nominal field of view.

\subsection{3D-Anchored Ground-Point Tracking}
\label{sec:method:anchors}

\subsubsection{Seeding, tracking, lifting.}
Shi--Tomasi corners~\cite{shi1994good} are detected over the full 4K frame; placement points are sampled on the placeable mask by farthest-point selection. Each point receives $K{=}11$ anchors from the corner set under semantic gating plus image-distance and depth bands that keep the constellation compact enough for local rigidity; a reserve pool of $60$ gated corners per point is tracked for replacement. All queries are propagated with pyramidal Lucas--Kanade flow~\cite{lucas1981iterative} and lifted to the camera frame by inverse projection $\pi^{-1}(\mathbf{x},z)$, with per-query depth smoothed by an exponential moving average -- the main stabilizer of the 3D constellation.

\subsubsection{Frame-invariant barycentric anchoring.}
The algorithm rests on a simple but non-trivial property: a world-static point is an \emph{affine} combination of world-static anchors with weights constant across frames.

\begin{property}[Frame invariance of affine anchor combinations]
\label{prop:invariance}
Let $\mathbf{p} \in \mathbb{R}^3$ and anchors $\{\mathbf{a}_k\}_{k=1}^{K}$, $K \ge 4$, be static in the world frame, with $\mathbf{p} = \sum_{k} w_k\,\mathbf{a}_k$, $\sum_{k} w_k = 1$. Then the same weights hold in any rigidly transformed system: $\mathbf{p}^{(t)} = \sum_k w_k\,\mathbf{a}_k^{(t)}$ for every camera frame $t$.
\end{property}

The proof is one line -- $\sum_k w_k (R\,\mathbf{a}_k + T) = R\,\mathbf{p} + T\sum_k w_k$, the crucial step being $\sum_k w_k = 1$ (see \cref{sec:supp:proof} for the full argument). The anchors are \emph{observed} every frame and the weight vector persists as the placement point's identity; with $K \gg 4$ anchors and several frames the fit is overconstrained, averaging out per-anchor depth noise.

\subsubsection{Constrained prediction.}
After an initialization window of $M{=}10$ frames we solve, for each placement,
\begin{equation}
\min_{\mathbf{w}} \; \sum_{m \in \mathcal{M}} \Big\lVert \sum_{k} w_k\, \mathbf{a}_k^{(m)} - \mathbf{p}^{(m)} \Big\rVert_2^2
+ \lambda^2 \Big\lVert \mathbf{w} - \tfrac{1}{K}\mathbf{1} \Big\rVert_2^2
\;\; \text{s.t.} \; \textstyle\sum_k w_k = 1,
\label{eq:baryfit}
\end{equation}
a stacked least-squares system with the constraint as a penalty row and a Tikhonov bias ($\lambda{=}0.5$, doubled while $\max_k|w_k| > 0.5$) toward the uniform barycentre -- without it the system is ill-conditioned and amplifies depth noise. Thereafter the placement point is \emph{predicted} at every frame,
\begin{equation}
\hat{\mathbf{p}}_t = \sum_{k} w_k\, \mathbf{a}_k^{(t)}, \qquad \hat{\mathbf{x}}_t = \pi(\hat{\mathbf{p}}_t),
\label{eq:predict}
\end{equation}
and by Property~\ref{prop:invariance} the weights need no update as the camera moves. A dead anchor (lost, near border, stalled) is replaced by the nearest healthy pool track and \cref{eq:baryfit} refit; when fewer than half the placements survive, a new \emph{epoch} begins with fresh corners and masks.

\subsection{Ground Rotation via an Anchor-Constellation SVD Fit}
\label{sec:method:rotation}

A person composited at frame $t_0$ and viewed at frame $t$ has rotated, in the camera frame, by the rigid rotation of the local ground patch -- for a drone, dominated by yaw. Over the anchor slots $\mathcal{K}_t$ whose \emph{identity} is unchanged since the reference frame, we fit
$R_t = \arg\min_{R \in SO(3)} \sum_{k \in \mathcal{K}_t}
\lVert R\,(\mathbf{a}_k^{\text{ref}} - \bar{\mathbf{a}}^{\text{ref}}) - (\mathbf{a}_k^{(t)} - \bar{\mathbf{a}}^{(t)}) \rVert_2^2$
by SVD with reflection guard~\cite{kabsch1976solution,umeyama1991least}, with frame-to-frame fallback and reference rebasing when correspondence degrades. The yaw about the optical axis, $\psi_t = \operatorname{atan2}((R_t)_{21}, (R_t)_{11})$, drives re-rendering: a placed person stays frontal \emph{with respect to the scene}, not the sweeping image frame.

\smallskip
\noindent\textbf{Why not direct point tracking?} The target is textureless by construction, whereas \cref{eq:predict} needs no appearance at the point: the median raw corner track on our panning 4K scenes survives ${\sim}35$ frames, anchored prediction sustains the full ${\sim}96$-frame span, and trackers~\cite{doersch2023tapir,karaev2024cotracker,xiao2024spatialtracker} output trajectories, not the metric depth ($h_{\text{px}} = f h_{\text{m}} / \tilde z_t$) and rotation $\psi_t$ that compositing needs -- both supplied in closed form at negligible cost.

\subsection{Avatar Rendering, Compositing and Verification}
\label{sec:method:render}

Each placement is assigned one of the ten communicative intents of~\cite{costea2026nonverbalrealtimehumanaiinteraction} and one of three sources: (i) \real{} -- chroma-extracted cut-outs of real actors; (ii) \realsynth{} -- the same performances captured as SMPL-X~\cite{pavlakos2019smplx} motion, retargeted onto rigged photorealistic characters and rendered at the placement's yaw $\psi_t$; (iii) \kimodo{} -- fully synthetic text-conditioned motion~\cite{rempe2026kimodo}, rendered identically. Sources share scenes, distances and compositing, so downstream differences isolate motion origin and appearance. Placement passes select maximal subsets of viable tracks (contiguous $80$-frame runs at $20$--$100$\,m) whose projected boxes never overlap; avatars are alpha-composited at 4K with scale from \cref{eq:predict}'s depth, the composite is swept by tiled YOLO11x-pose~\cite{jocher2024yolo11}, and a placement is \emph{kept} only if detected in $\ge 80$ consecutive frames, harvested as a $(80,17,2)$ keypoint tensor with intent, source, scene and median metric distance.

%-------------------------------------------------------------------------
\section{The Drones2BodyLanguage Dataset}
\label{sec:dataset}

We ran the pipeline over eight Romanian drone scenes. \Cref{tab:dataset} summarizes the corpus: $3{,}580$ placements balanced across sources and intents at $32.7$--$98.5$\,m; splits follow the 10-intent benchmark format of~\cite{costea2026nonverbalrealtimehumanaiinteraction} and are leakage-free (no shared motion identities or scenes; \textsc{jupiter}, \textsc{olanesti} held out). Over $3{,}760$ logged attempts (${\sim}404$k frames), per-frame detection is essentially perfect to $70$\,m and degrades to $0.942$ at $90$--$100$\,m; rendered avatars hold ${\ge}0.98$ throughout while \real{} cut-outs, whose matting inherits video noise, fall to $0.848$; keypoint visibility is flat in distance but source-dependent ($16.4$/$16.0$/$13.7$ of $17$).

\begin{table}[tb]
  \caption{The corpus: $464$--$546$ placements per main scene, $73$ for the high-altitude \textsc{targoviste} pilot; each is a detection-verified $80$-frame window of $17$ COCO keypoints with intent, source, scene and metric distance}
  \label{tab:dataset}
  \centering
  \footnotesize
  \setlength{\tabcolsep}{5pt}
  \begin{tabular}{@{}lrrrr@{}}
    \toprule
    & \real & \realsynth & \kimodo & total \\
    \midrule
    placements & $1{,}167$ & $1{,}192$ & $1{,}221$ & $3{,}580$ \\
    per intent (min--max) & $98$--$131$ & $113$--$124$ & $115$--$129$ & $341$--$369$ \\
    \bottomrule
  \end{tabular}
\end{table}

%-------------------------------------------------------------------------
\section{Experiments}
\label{sec:experiments}

We evaluate (i) keypoint fidelity through the pipeline, (ii) zero-shot transfer of close-range models, (iii) placement-vs-no-placement training across twelve architectures, and (iv) ablations of rotation, source mixing and finetuning order.

\subsection{Keypoint Fidelity: Rotation Dominates}
\label{sec:exp:l2}

We measure torso-normalized detector error ($L_{2n}$) against ground-truth keypoints (animation-rig joints projected through the render camera) along the pipeline. Frontal full-resolution renders score a median of $0.124$; rotating the same performances to $\pm75^\circ$ yaw -- exactly what $\psi_t$ induces at placement -- raises it to $0.150$ ($+21\%$), while compositing into 4K scenes and re-detecting at placement scale adds little further error. Viewpoint rotation, not compositing or downscaling, is the dominant fidelity cost.

\subsection{Zero-Shot Transfer of Close-Range Models}
\label{sec:exp:crossdomain}

Frozen Transformer classifiers trained on the close-range benchmark~\cite{costea2026nonverbalrealtimehumanaiinteraction}, evaluated on $947$ detection-verified placement windows at $60$--$131$\,m, show: \textbf{(1)}~the source gap is a \emph{motion} gap -- \real{} ($0.51$--$0.58$) $>$ \realsynth{} ($0.32$--$0.40$) $\gg$ \kimodo{} (chance), although \realsynth{} carries real motion under synthetic appearance; \textbf{(2)}~the \kimodo{} collapse is distributional -- the frozen model assigns $315$ of $325$ \kimodo{} windows to one class, yet a nearest-centroid probe separates its actions at $0.686$; \textbf{(3)}~accuracy is flat from $60$ to $130$\,m: distance costs \emph{detections}, not classification-given-detection.

\subsection{Placement vs.\ No-Placement Training}
\label{sec:exp:placement}

Does training on placed data close this gap? We train the twelve skeleton-sequence baselines of~\cite{costea2026nonverbalrealtimehumanaiinteraction} -- eight dual-task models and four classification-only graph networks~\cite{chen2021ctrgcn,yan2018stgcn} -- under the benchmark's fixed protocol ($60$ input frames $\to$ forecast $30$ $+$ classify $10$ intents), on leakage-free splits (no shared motion identities or scenes; ${\sim}950$--$999$ train / $250$ test clips per source); \cref{tab:placement} aggregates the three regimes (per-model results in \cref{sec:supp:full}).

\begin{table}[tb]
  \caption{Placement vs.\ no-placement training on scene- and motion-disjoint splits, aggregated over the $12$ baselines (chance $=0.10$; MAE over the $8$ dual-task models). Zero-shot $<$ from-scratch $<$ finetune on every source. Different stop epochs (ep) were tested}
  \label{tab:placement}
  \centering
  \footnotesize
  \setlength{\tabcolsep}{3.2pt}
  \resizebox{\linewidth}{!}{%
  \begin{tabular}{@{}lrrrrrrrrr@{}}
    \toprule
    & \multicolumn{3}{c}{\real} & \multicolumn{3}{c}{\realsynth} & \multicolumn{3}{c}{\kimodo} \\
    \cmidrule(lr){2-4}\cmidrule(lr){5-7}\cmidrule(l){8-10}
    Training regime & mean & best & MAE & mean & best & MAE & mean & best & MAE \\
    \midrule
    (a) close-range only, zero-shot   & $0.476$ & $0.617$ & $0.060$ & $0.319$ & $0.492$ & $0.074$ & $0.199$ & $0.252$ & $0.055$ \\
    (b) placement, from scratch, $200$ ep & $0.675$ & $0.835$ & $0.058$ & $0.580$ & $0.741$ & $0.067$ & $0.486$ & $0.607$ & $0.043$ \\
    (c) close-range${\to}$placement, $100$ ep & $\mathbf{0.744}$ & $\mathbf{0.864}$ & $\mathbf{0.054}$ & $\mathbf{0.671}$ & $\mathbf{0.789}$ & $\mathbf{0.061}$ & $\mathbf{0.577}$ & $\mathbf{0.704}$ & $\mathbf{0.040}$ \\
    \bottomrule
  \end{tabular}}
\end{table}

\noindent Mean accuracy rises from the zero-shot floor to $0.744$/$0.671$/$0.577$ finetuned (best $0.864$ on \real); finetuning beats from-scratch on $34$ of $36$ model--source pairs with half the epochs, and $\real > \realsynth > \kimodo$ holds within each regime; metrics are window-level, the test split doubling as validation.

\subsection{Ablations: Rotation, Source Mixing, Finetuning Order}
\label{sec:exp:ablation}

Replaying every \realsynth{} placement at constant zero yaw -- same tracks, motions, avatars and scenes -- and retraining all twelve models shifts mean accuracy by at most $0.003$ in every regime (\cref{tab:ablation}, left): the geometric realism of \cref{sec:method:rotation} costs nothing in learnability. Regenerating the \kimodo{} motion inventory under the same protocol, in contrast, moves the checkpoint-identical zero-shot probe by $+10.8$ points: \emph{which} motions the data contains matters, \emph{how} the avatar is oriented does not.

Mixing justifies \realsynth{} (\cref{tab:ablation}, right): from scratch, \real$+$\realsynth{} beats single-source training on both component test sets ($+3.2$ on \real, $+7.8$ on \realsynth), and the two-hop finetune close-range${\to}$\realsynth${\to}$\real{} beats the direct recipe ($0.757$ vs $0.744$ mean), the synthetic hop pre-adapting the representation to placement rendering.

\begin{table}[tb]
  \caption{Ablations over the $12$ baselines. Left: mean accuracy on the \realsynth{} test set with/without the placement rotation. Right: mean/best accuracy on the \real{} test set. The \realsynth versions yield the best results.}
  \label{tab:ablation}
  \centering
  \scriptsize
  \begin{minipage}[t]{0.20\linewidth}
    \centering
    \setlength{\tabcolsep}{2pt}
    \begin{tabular}{@{}lrrr@{}}
      \toprule
      regime & \textsc{rot} mAcc & \textsc{norot} mAcc & $\Delta$ \\
      \midrule
      zero-shot & $0.319$ & $0.318$ & $-0.001$ \\
      from scratch & $0.580$ & $0.579$ & $-0.001$ \\
      finetuned & $0.671$ & $0.668$ & $-0.003$ \\
      \bottomrule
    \end{tabular}
  \end{minipage}\hfill
  \begin{minipage}[t]{0.54\linewidth}
    \centering
    \setlength{\tabcolsep}{4pt}
    \begin{tabular}{@{}lrr@{}}
      \toprule
      training set & mean & best \\
      \midrule
      \real, from scratch & $0.675$ & $0.835$ \\
      \real$+$\realsynth, from scratch & $0.707$ & $0.866$ \\
      \real, finetuned & $0.744$ & $0.864$ \\
      \real$+$\realsynth, finetuned & $0.735$ & $\mathbf{0.888}$ \\
      chain \realsynth${\to}$\real, finetuned & $\mathbf{0.757}$ & $0.887$ \\
      \bottomrule
    \end{tabular}
  \end{minipage}
\end{table}

\subsection{Generalization to Real Countryside and Resort Footage}
\label{sec:exp:raciu}

\begin{figure}[tb]
  \centering
  \includegraphics[width=\linewidth]{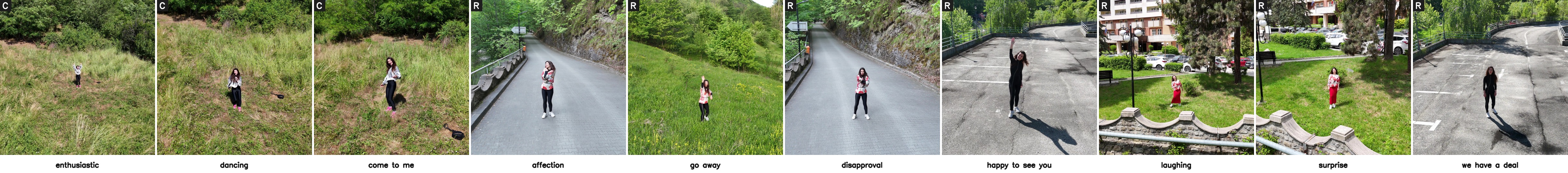}
  \caption{Samples from the two real validation scenes, all ten intents: person-centred crops from the Countryside (C, first three) and Resort (R) from 4K footage}
  \label{fig:real-scenes}
\end{figure}

As a final probe, we recorded two un-composited validation sets, one actor performing the ten intents at 4K (\cref{fig:real-scenes}): \emph{Countryside} ($168$ detection-verified $80$-frame clips) and the longer-range \emph{Resort} ($62$ clips, nine classes). Detection transfers readily (\cref{tab:real-scenes-zeroshot}). Evaluated inference-only (\cref{tab:raciu}), the mixed-source models rank first on both scenes ($0.426$ and $0.407$ mean for the \real$+$\realsynth{} finetune), and mixed from-scratch training nearly substitutes for real-data pretraining ($0.405$/$0.406$). In-domain ranking does not predict new-scene ranking.

\begin{table}[tb]
  \begin{minipage}[t]{0.38\linewidth}
    \caption{Zero-shot transfer to the two real scenes (chance $=0.10$), tiled YOLO11x-pose detection; accuracy is the zero-shot Transformer under the per-sequence protocol}
    \label{tab:real-scenes-zeroshot}
    \centering
    \scriptsize
    \setlength{\tabcolsep}{3pt}
    \begin{tabular}{@{}lrr@{}}
      \toprule
      & Countrys. & Resort \\
      \midrule
      depth (m) & $3$--$30$ & $5$--$55$ \\
      det.\ rate & $0.920$ & $0.922$ \\
      zero-shot acc. & $0.284$ & $0.222$ \\
      \bottomrule
    \end{tabular}
  \end{minipage}\hfill
  \begin{minipage}[t]{0.58\linewidth}
    \caption{Real scene validation (unseen, no placement) over the $12$ architectures, Countryside and Resort; the combination of real and \realsynth yields the best results. Mixed (\textsc{rs} $=$ \realsynth) rows use the \textsc{rot} variant.}
    \label{tab:raciu}
    \centering
    \scriptsize
    \setlength{\tabcolsep}{3pt}
    \begin{tabular}{@{}llrrrr@{}}
      \toprule
      & & \multicolumn{2}{c}{Countryside} & \multicolumn{2}{c@{}}{Resort} \\
      \cmidrule(lr){3-4} \cmidrule(l){5-6}
      Regime & source & mAcc & bAcc & mAcc & bAcc \\
      \midrule
      (a) zero-shot & \real{} & $0.356$ & $0.417$ & $0.348$ & $0.464$ \\
      \midrule
      \multirow{4}{*}{\shortstack[l]{(b) from\\scratch}} & \real{} & $0.347$ & $0.465$ & $0.338$ & $0.427$ \\
       & \realsynth{} & $0.311$ & $0.399$ & $0.255$ & $0.398$ \\
       & \kimodo{} & $0.086$ & $0.126$ & $0.117$ & $0.188$ \\
       & \real$+$\textsc{rs} & $0.405$ & $0.547$ & $0.406$ & $\mathbf{0.567}$ \\
      \midrule
      \multirow{4}{*}{(c) finetuned} & \real{} & $0.421$ & $0.526$ & $0.387$ & $0.494$ \\
       & \realsynth{} & $0.337$ & $0.441$ & $0.340$ & $0.475$ \\
       & \kimodo{} & $0.152$ & $0.296$ & $0.195$ & $0.355$ \\
       & \real$+$\textsc{rs} & $\mathbf{0.426}$ & $\mathbf{0.555}$ & $\mathbf{0.407}$ & $0.514$ \\
      \bottomrule
    \end{tabular}
    \vspace{-5mm}
  \end{minipage}
\end{table}

%-------------------------------------------------------------------------
\section{Conclusions}
\label{sec:conclusions}

We presented \emph{Drones2BodyLanguage} and its placing algorithm -- a geometric world model with frame-invariant affine anchoring and SVD-fitted ground rotation. Placement training lifts twelve architectures from $0.48$/$0.32$/$0.20$ zero-shot accuracy to $0.74$/$0.67$/$0.58$; mixing real and retargeted placements pushes further ($0.757$ with the synthetic-first chain) and ranks first on two new scenes. \emph{Limitations:} locally rigid ground, the real character cutouts are not re-renderable without the 3D pipeline, more camera angles and placement scenarios could improve robustness.

% ---- Bibliography ----
\bibliographystyle{splncs04}
\bibliography{main}

%=========================================================================
% APPENDIX (formerly the supplementary material)
%=========================================================================
\clearpage
\setcounter{section}{0}
\renewcommand{\thesection}{\Alph{section}}
\renewcommand{\thetable}{\Alph{section}\arabic{table}}
\renewcommand{\thefigure}{\Alph{section}\arabic{figure}}
\setcounter{table}{0}
\setcounter{figure}{0}

\section*{Appendix}

\noindent\sloppy This appendix provides (\cref{sec:supp:proof}) the full proof and discussion of the frame-invariance property that underpins the placement tracker; (\cref{sec:supp:full}) the complete per-model results behind the aggregated numbers in \cref{sec:exp:placement}; (\cref{sec:supp:detrate}) the full detection-rate-versus-distance breakdown; (\cref{sec:supp:raciu}) the full per-model breakdown and qualitative material for the countryside validation dataset of \cref{sec:exp:raciu}; and (\cref{sec:supp:qual}) further qualitative material.

%-------------------------------------------------------------------------
\section{Frame Invariance of Affine Anchor Combinations}
\label{sec:supp:proof}

We restate Property~\ref{prop:invariance} and give the full argument that was compressed in the main text.

\begin{property}[Frame invariance]
Let $\mathbf{p} \in \mathbb{R}^3$ be a scene point that is static in the world frame, and let $\{\mathbf{a}_k\}_{k=1}^{K}$, $K \ge 4$, be static anchor points, all expressed in a common coordinate system, such that $\mathbf{p} = \sum_{k=1}^{K} w_k\,\mathbf{a}_k$ with $\sum_{k=1}^{K} w_k = 1$. Then the same weights reproduce the point in any rigidly transformed coordinate system: for every camera frame $t$, $\mathbf{p}^{(t)} = \sum_k w_k\,\mathbf{a}_k^{(t)}$.
\end{property}

\begin{proof}
Let the camera pose at frame $t$ induce the rigid transform $\mathbf{x} \mapsto R_t\,\mathbf{x} + T_t$ from the world frame to the camera frame, with $R_t \in SO(3)$, $T_t \in \mathbb{R}^3$. Then
\begin{align}
\sum_k w_k\,\mathbf{a}_k^{(t)}
= \sum_k w_k\,\big(R_t\,\mathbf{a}_k + T_t\big)
&= R_t \Big(\sum_k w_k\,\mathbf{a}_k\Big) + \Big(\sum_k w_k\Big) T_t \nonumber\\
&= R_t\,\mathbf{p} + T_t = \mathbf{p}^{(t)} .
\end{align}
The rotation part commutes with \emph{any} linear combination; the affine constraint $\sum_k w_k = 1$ is what makes the translation part factor correctly. \qed
\end{proof}

\subsubsection{Why the constraint is necessary.}
Without $\sum_k w_k = 1$ the translation term becomes $(\sum_k w_k)\,T_t \ne T_t$, so a plain linear combination is only preserved under pure rotations and fails as soon as the drone translates. In homogeneous coordinates the statement is immediate: appending $1$ to each point, $\tilde{\mathbf{p}} = \sum_k w_k\,\tilde{\mathbf{a}}_k$ requires the fourth coordinate to satisfy $1 = \sum_k w_k$, and every rigid (indeed, every affine) map acts linearly on homogeneous coordinates. The property is exact -- no small-motion or planarity approximation is involved -- and it holds for similarity transforms as well, so a global scale ambiguity in monocular depth does not break it either.

\subsubsection{Existence and conditioning.}
For $K \ge 4$ anchors in general position (not all coplanar), the affine hull of $\{\mathbf{a}_k\}$ is all of $\mathbb{R}^3$, so weights reproducing any $\mathbf{p}$ exist; for $K > 4$ they form a $(K{-}4)$-dimensional affine family. The regularized least-squares fit of \cref{eq:baryfit} selects, among all consistent weight vectors, the one closest to uniform $\tfrac{1}{K}\mathbf{1}$: this spreads the prediction over the whole constellation, so no single noisy anchor dominates, and averages the per-frame depth noise over $K$ anchors and $M$ frames. The depth band used at seeding ($z_p/\rho < z(\mathbf{a}_k) < \rho\,z_p$, excluding the point's own depth shell) exists precisely to avoid the degenerate all-coplanar case while keeping the constellation compact enough that the static-world assumption holds for all its members.

\subsubsection{What breaks it in practice.}
The two error sources are (i) depth noise, which perturbs the lifted anchors $\mathbf{a}_k^{(t)}$ -- damped by the EMA and by the redundancy of the fit -- and (ii) anchor drift, where a Lucas--Kanade track slides off its world point, violating the ``static anchor'' premise for that $k$; this is handled by the stall/border/loss detectors and pool replacement of \cref{sec:method:anchors}, followed by a refit.

%-------------------------------------------------------------------------
\section{Full Per-Model Results on the Scene-Disjoint Placement Splits}
\label{sec:supp:full}

\begin{figure}[t]
  \centering
  \includegraphics[width=0.78\linewidth]{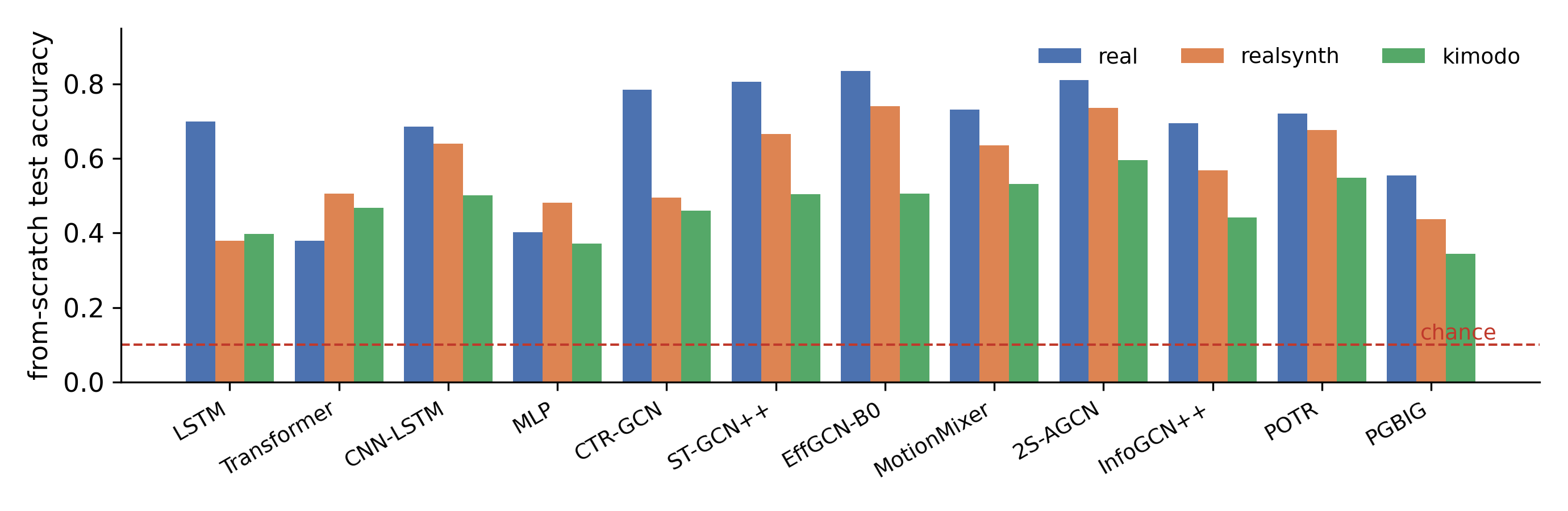}\\[2pt]
  \includegraphics[width=0.78\linewidth]{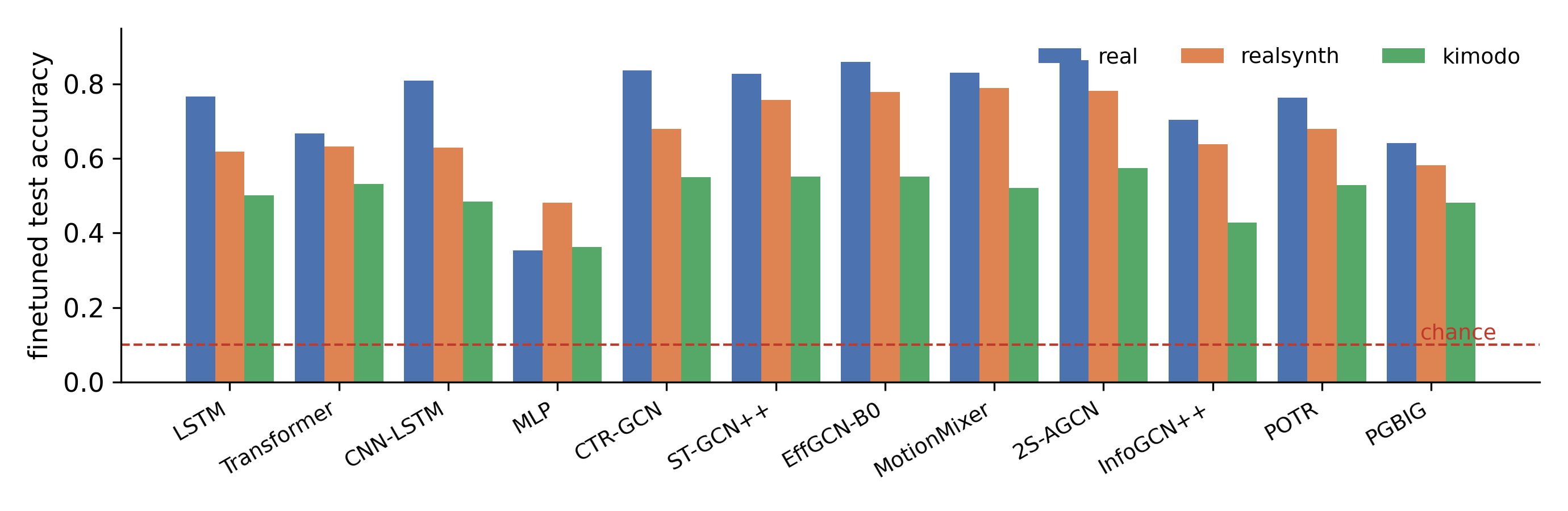}
  \caption{Per-model test accuracy on the placement splits: trained from scratch (top) and finetuned from close-range checkpoints (bottom).}
  \label{fig:supp:permodel}
\end{figure}

\begin{figure}[t]
  \centering
  \includegraphics[width=0.92\linewidth]{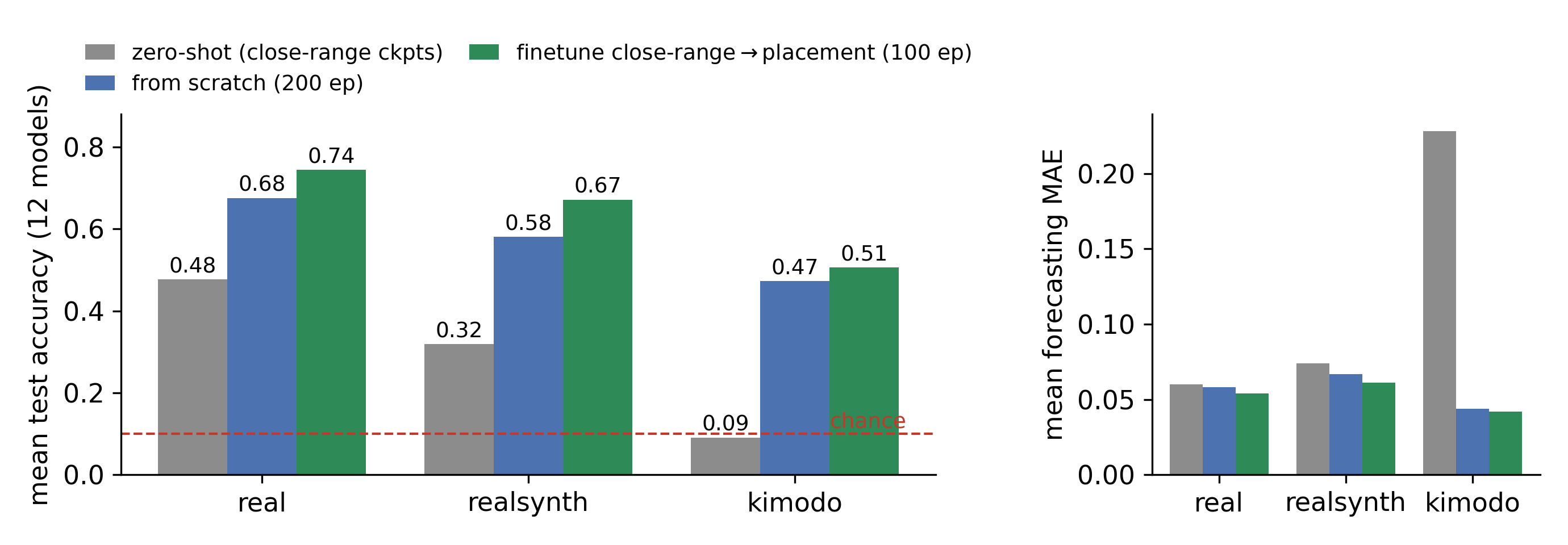}
  \caption{Mean test accuracy over the twelve architectures per regime and placement source (left) and mean forecasting MAE of the eight dual-task models (right). Every source improves monotonically from zero-shot to scratch to finetuned; \kimodo{} forecasting collapses only in the zero-shot regime.}
  \label{fig:supp:regimes}
\end{figure}

\Cref{tab:supp:placement} expands the aggregated \cref{tab:placement}: test accuracy of every architecture under the three training regimes, per placement source, on the leakage-free splits (train and test share no motion identities and no scenes; \textsc{jupiter}, \textsc{olanesti} held out; chance $0.10$). \Cref{fig:supp:permodel} plots the per-model accuracy for the from-scratch and finetuned regimes, and \cref{fig:supp:regimes} summarizes all three regimes side by side, together with the mean forecasting MAE of the eight dual-task models.

\begin{table}[t]
  \caption{Per-model test accuracy on the scene- and motion-disjoint placement splits (chance $0.10$). Regimes: (a) close-range checkpoints, zero-shot; (b) trained from scratch on placement data ($200$ epochs); (c) close-range checkpoints finetuned on placement data ($100$ epochs). \textbf{Bold} = best per column.}
  \label{tab:supp:placement}
  \centering
  \scriptsize
  \setlength{\tabcolsep}{3.0pt}
  \begin{tabular}{@{}lrrrrrrrrr@{}}
    \toprule
    & \multicolumn{3}{c}{(a) zero-shot} & \multicolumn{3}{c}{(b) from scratch} & \multicolumn{3}{c}{(c) finetuned} \\
    \cmidrule(lr){2-4}\cmidrule(lr){5-7}\cmidrule(l){8-10}
    Model & \real & \realsynth & \kimodo & \real & \realsynth & \kimodo & \real & \realsynth & \kimodo \\
    \midrule
    LSTM        & $0.410$ & $0.381$ & $0.063$ & $0.699$ & $0.379$ & $0.397$ & $0.767$ & $0.618$ & $0.501$ \\
    Transformer & $0.336$ & $0.314$ & $0.079$ & $0.380$ & $0.506$ & $0.468$ & $0.667$ & $0.633$ & $0.531$ \\
    CNN-LSTM    & $0.496$ & $0.292$ & $0.058$ & $0.685$ & $0.640$ & $0.501$ & $0.809$ & $0.629$ & $0.484$ \\
    MLP         & $0.332$ & $0.274$ & $0.096$ & $0.403$ & $0.481$ & $0.372$ & $0.354$ & $0.481$ & $0.363$ \\
    CTR-GCN     & $0.596$ & $0.287$ & $0.094$ & $0.784$ & $0.495$ & $0.460$ & $0.837$ & $0.680$ & $0.550$ \\
    ST-GCN++    & $0.544$ & $0.272$ & $0.092$ & $0.806$ & $0.666$ & $0.505$ & $0.828$ & $0.757$ & $0.551$ \\
    EffGCN-B0   & $0.563$ & $0.288$ & $0.095$ & $\mathbf{0.835}$ & $\mathbf{0.741}$ & $0.506$ & $0.860$ & $0.779$ & $0.552$ \\
    MotionMixer & $\mathbf{0.617}$ & $\mathbf{0.492}$ & $0.099$ & $0.732$ & $0.635$ & $0.531$ & $0.830$ & $\mathbf{0.789}$ & $0.521$ \\
    2S-AGCN     & $0.488$ & $0.267$ & $\mathbf{0.140}$ & $0.810$ & $0.736$ & $\mathbf{0.596}$ & $\mathbf{0.864}$ & $0.782$ & $\mathbf{0.575}$ \\
    InfoGCN++   & $0.404$ & $0.347$ & $0.048$ & $0.695$ & $0.569$ & $0.442$ & $0.704$ & $0.639$ & $0.428$ \\
    POTR        & $0.540$ & $0.384$ & $0.123$ & $0.720$ & $0.676$ & $0.548$ & $0.764$ & $0.680$ & $0.528$ \\
    PGBIG       & $0.393$ & $0.227$ & $0.100$ & $0.554$ & $0.438$ & $0.344$ & $0.642$ & $0.582$ & $0.481$ \\
    \midrule
    mean        & $0.477$ & $0.319$ & $0.091$ & $0.675$ & $0.580$ & $0.473$ & $0.744$ & $0.671$ & $0.505$ \\
    \bottomrule
  \end{tabular}
\end{table}

%-------------------------------------------------------------------------
\section{Detection Rate versus Distance: Full Per-Bin Table}
\label{sec:supp:detrate}

\Cref{tab:detrate_supp} gives the per-frame YOLO11x-pose detection probability of composited placements against true metric distance, over all $3{,}760$ logged placement attempts (${\sim}404$k composited frames) of the multi-scene corpus, per $10$\,m bin and per source. Detection is essentially perfect to $70$\,m and degrades monotonically to $0.942$ at $90$--$100$\,m; rendered avatars (\realsynth{}, \kimodo) hold ${\ge}0.98$ throughout, while \real{} cut-outs, whose matting inherits real video noise, fall to $0.848$ in the last bin.

\begin{table}[tb]
  \caption{Detection rate vs.\ distance over \emph{all} placement attempts (kept + short + undetected), per $10$\,m bin and per source. Multi-scene corpus, tiled YOLO11x-pose}
  \label{tab:detrate_supp}
  \centering
  \footnotesize
  \setlength{\tabcolsep}{4.5pt}
  \begin{tabular}{@{}lrrrrrrr@{}}
    \toprule
    dist (m) & comp.\ fr & det.\ fr & det.\ rate & placem. & \real & \realsynth & \kimodo \\
    \midrule
    30--50  & $15{,}360$  & $15{,}360$  & $1.000$ & $168$   & $1.000$ & $1.000$ & $1.000$ \\
    50--60  & $27{,}003$  & $26{,}975$  & $0.999$ & $283$   & $0.997$ & $1.000$ & $1.000$ \\
    60--70  & $59{,}442$  & $59{,}320$  & $0.998$ & $593$   & $0.994$ & $1.000$ & $1.000$ \\
    70--80  & $105{,}464$ & $104{,}769$ & $0.993$ & $1{,}002$ & $0.983$ & $0.999$ & $1.000$ \\
    80--90  & $120{,}288$ & $117{,}616$ & $0.978$ & $1{,}084$ & $0.943$ & $0.996$ & $0.997$ \\
    90--100 & $76{,}292$  & $71{,}848$  & $0.942$ & $630$   & $0.848$ & $0.984$ & $0.996$ \\
    \bottomrule
  \end{tabular}
\end{table}

\section{Countryside Validation Dataset: Full Breakdown}
\label{sec:supp:raciu}

\Cref{tab:raciu} of the main text reports per-source means and bests on the $168$ real clips of the held-out countryside scene ($4$K$@30$\,fps, one actor, all ten intents, inference only). \Cref{tab:supp:raciu-acc} gives the complete $84$-checkpoint accuracy matrix and \cref{tab:supp:raciu-mae} the forecasting MAE of the eight dual-task models. Macro-F1 tracks accuracy closely for all real-source models (mean $0.350$/$0.424$/$0.355$ for scratch-\real/finetuned-\real/zero-shot), confirming no single-class collapse.

\begin{figure}[t]
  \centering
  \includegraphics[width=0.94\linewidth]{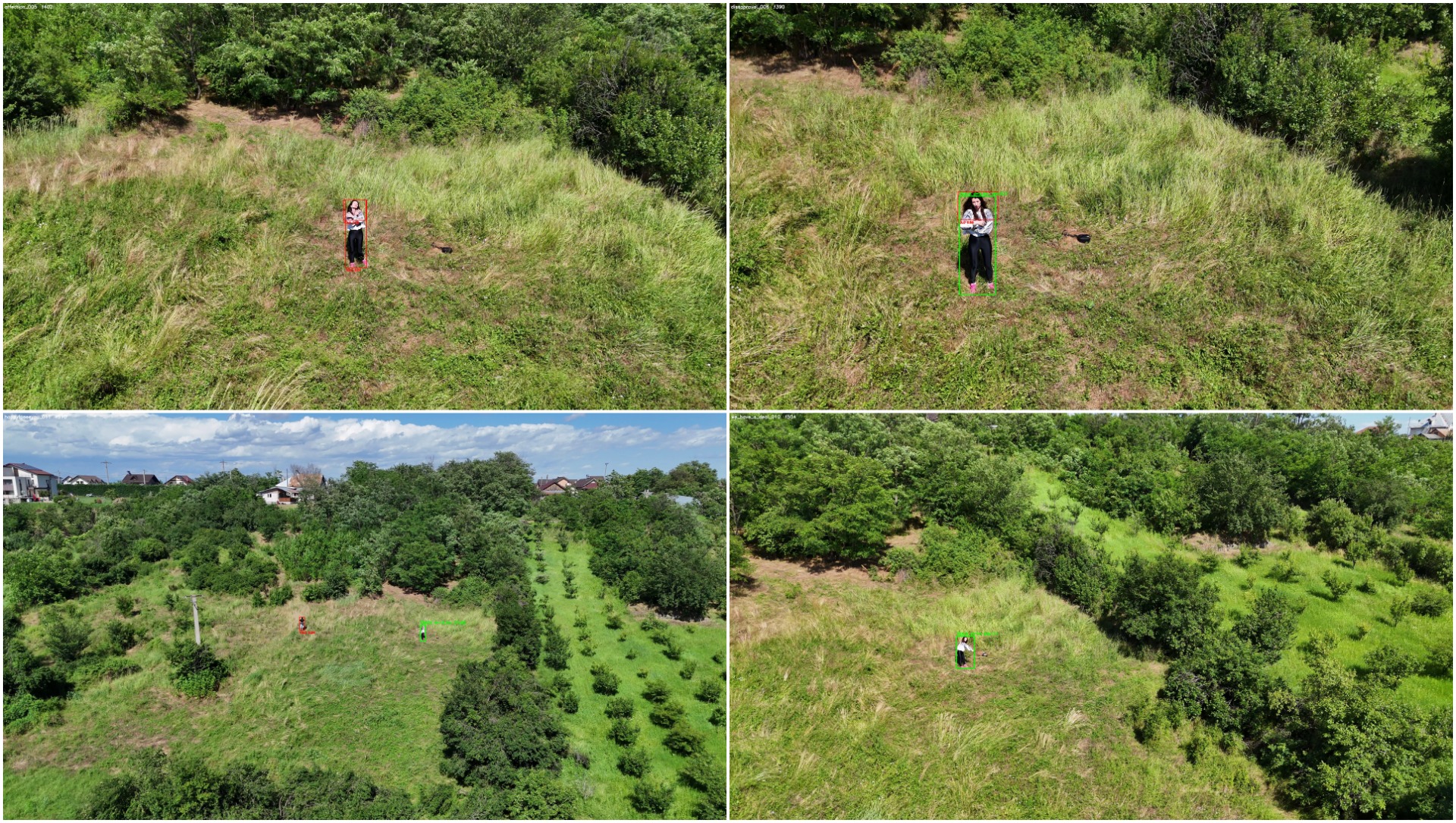}
  \caption{Tracking-pipeline output on four countryside validation clips (distinct intents). Green: the tracked central subject whose keypoints form the evaluation clips.}
  \label{fig:supp:raciu-montage}
\end{figure}

\begin{figure}[t]
  \centering
  \includegraphics[width=0.94\linewidth]{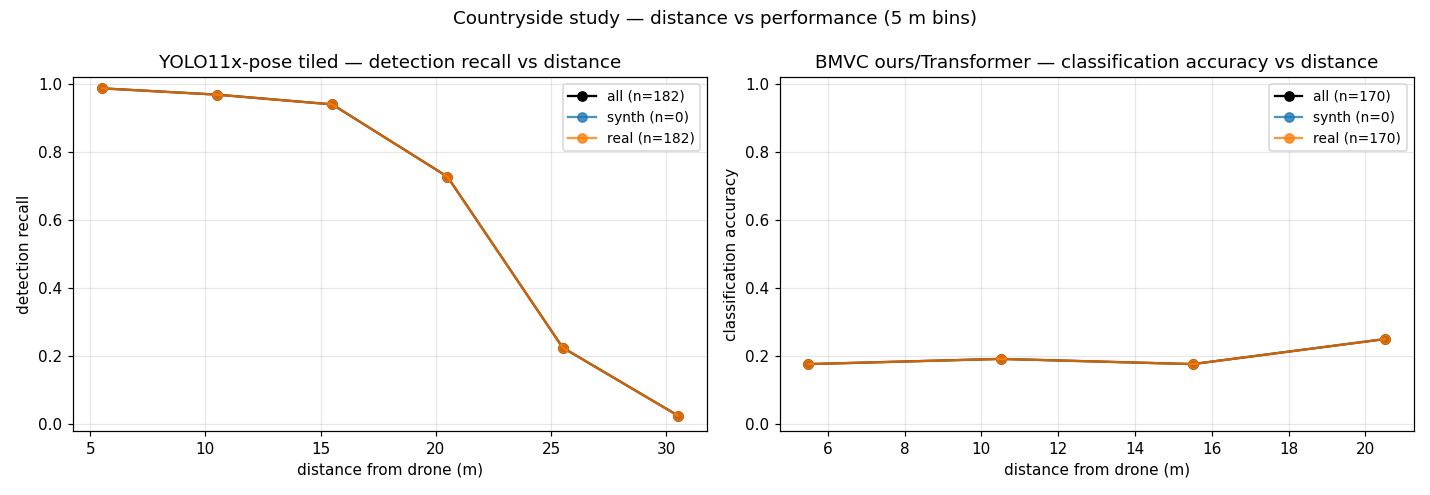}
  \caption{Countryside validation extraction quality vs.\ camera-to-subject distance ($5$\,m bins, $182$ windows): tiled YOLO11x-pose detection recall (left) and zero-shot Transformer classification accuracy (right). Recall stays above $0.94$ up to $15$\,m and degrades beyond; zero-shot accuracy is roughly flat over the covered range, consistent with the source-gap-dominates finding of the main text.}
  \label{fig:supp:raciu-dist}
\end{figure}

\Cref{fig:supp:raciu-montage} shows the tracking pipeline output on the countryside clips, and \cref{fig:supp:raciu-dist} the detection recall and zero-shot classification accuracy as a function of camera-to-subject distance for these clips ($5$--$30$\,m, closer than the placement corpus band).

\begin{table}[t]
  \caption{Test accuracy of all $84$ checkpoints on the countryside validation clips (chance $0.10$). Column groups as in \cref{tab:supp:placement}. \textbf{Bold} = best per column.}
  \label{tab:supp:raciu-acc}
  \centering
  \footnotesize
  \setlength{\tabcolsep}{2.6pt}
  \begin{tabular}{@{}lrrrrrrr@{}}
    \toprule
    & (a) zero-shot & \multicolumn{3}{c}{(b) from scratch} & \multicolumn{3}{c}{(c) finetuned} \\
    \cmidrule(lr){2-2}\cmidrule(lr){3-5}\cmidrule(l){6-8}
    Model & \real & \real & \realsynth & \kimodo & \real & \realsynth & \kimodo \\
    \midrule
    LSTM        & $0.386$ & $0.338$ & $0.298$ & $\mathbf{0.126}$ & $\mathbf{0.526}$ & $0.361$ & $\mathbf{0.296}$ \\
    Transformer & $0.284$ & $0.219$ & $0.353$ & $0.099$ & $0.441$ & $0.296$ & $0.109$ \\
    CNN-LSTM    & $0.352$ & $0.287$ & $0.313$ & $0.043$ & $0.406$ & $0.326$ & $0.063$ \\
    MLP         & $0.336$ & $0.219$ & $0.241$ & $0.053$ & $0.183$ & $0.201$ & $0.058$ \\
    CTR-GCN     & $0.349$ & $0.329$ & $0.221$ & $0.065$ & $0.417$ & $0.317$ & $0.173$ \\
    ST-GCN++    & $0.374$ & $0.416$ & $0.237$ & $0.116$ & $0.507$ & $0.365$ & $0.118$ \\
    EffGCN-B0   & $0.393$ & $0.395$ & $\mathbf{0.399}$ & $0.082$ & $0.500$ & $\mathbf{0.441}$ & $0.219$ \\
    MotionMixer & $\mathbf{0.417}$ & $\mathbf{0.465}$ & $0.298$ & $0.086$ & $0.443$ & $0.410$ & $0.219$ \\
    2S-AGCN     & $0.345$ & $0.381$ & $0.349$ & $0.094$ & $0.464$ & $0.330$ & $0.123$ \\
    InfoGCN++   & $0.381$ & $0.426$ & $0.344$ & $0.119$ & $0.456$ & $0.395$ & $0.166$ \\
    POTR        & $0.384$ & $0.340$ & $0.291$ & $0.046$ & $0.400$ & $0.351$ & $0.207$ \\
    PGBIG       & $0.269$ & $0.344$ & $0.392$ & $0.103$ & $0.312$ & $0.254$ & $0.071$ \\
    \midrule
    mean        & $0.356$ & $0.347$ & $0.311$ & $0.086$ & $0.421$ & $0.337$ & $0.152$ \\
    \bottomrule
  \end{tabular}
\end{table}

\begin{table}[t]
  \caption{Forecasting MAE on the countryside validation dataset for the eight dual-task models (lower is better). Real-source MAE transfers to the new scene nearly intact, while \kimodo-trained forecasting degrades $2$--$4\times$.}
  \label{tab:supp:raciu-mae}
  \centering
  \footnotesize
  \setlength{\tabcolsep}{2.6pt}
  \begin{tabular}{@{}lrrrrrrr@{}}
    \toprule
    & (a) zero-shot & \multicolumn{3}{c}{(b) from scratch} & \multicolumn{3}{c}{(c) finetuned} \\
    \cmidrule(lr){2-2}\cmidrule(lr){3-5}\cmidrule(l){6-8}
    Model & \real & \real & \realsynth & \kimodo & \real & \realsynth & \kimodo \\
    \midrule
    LSTM        & $0.078$ & $0.078$ & $0.092$ & $0.177$ & $0.076$ & $0.086$ & $0.160$ \\
    Transformer & $0.080$ & $0.086$ & $0.090$ & $0.184$ & $0.077$ & $0.080$ & $0.167$ \\
    CNN-LSTM    & $0.087$ & $0.085$ & $0.099$ & $0.176$ & $0.082$ & $0.090$ & $0.174$ \\
    MLP         & $0.101$ & $0.145$ & $0.123$ & $0.637$ & $0.097$ & $0.103$ & $0.342$ \\
    MotionMixer & $0.077$ & $0.073$ & $0.093$ & $0.149$ & $0.075$ & $0.079$ & $0.172$ \\
    InfoGCN++   & $\mathbf{0.071}$ & $\mathbf{0.068}$ & $\mathbf{0.072}$ & $0.172$ & $\mathbf{0.068}$ & $\mathbf{0.073}$ & $\mathbf{0.117}$ \\
    POTR        & $0.078$ & $0.081$ & $0.083$ & $\mathbf{0.118}$ & $0.075$ & $0.080$ & $0.133$ \\
    PGBIG       & $0.084$ & $0.084$ & $0.089$ & $0.163$ & $0.084$ & $0.088$ & $0.146$ \\
    \midrule
    mean        & $0.082$ & $0.088$ & $0.093$ & $0.222$ & $0.079$ & $0.085$ & $0.177$ \\
    \bottomrule
  \end{tabular}
\end{table}

%-------------------------------------------------------------------------
\section{Additional Qualitative Material}
\label{sec:supp:qual}

\begin{figure}[t]
  \centering
  \includegraphics[width=0.86\linewidth]{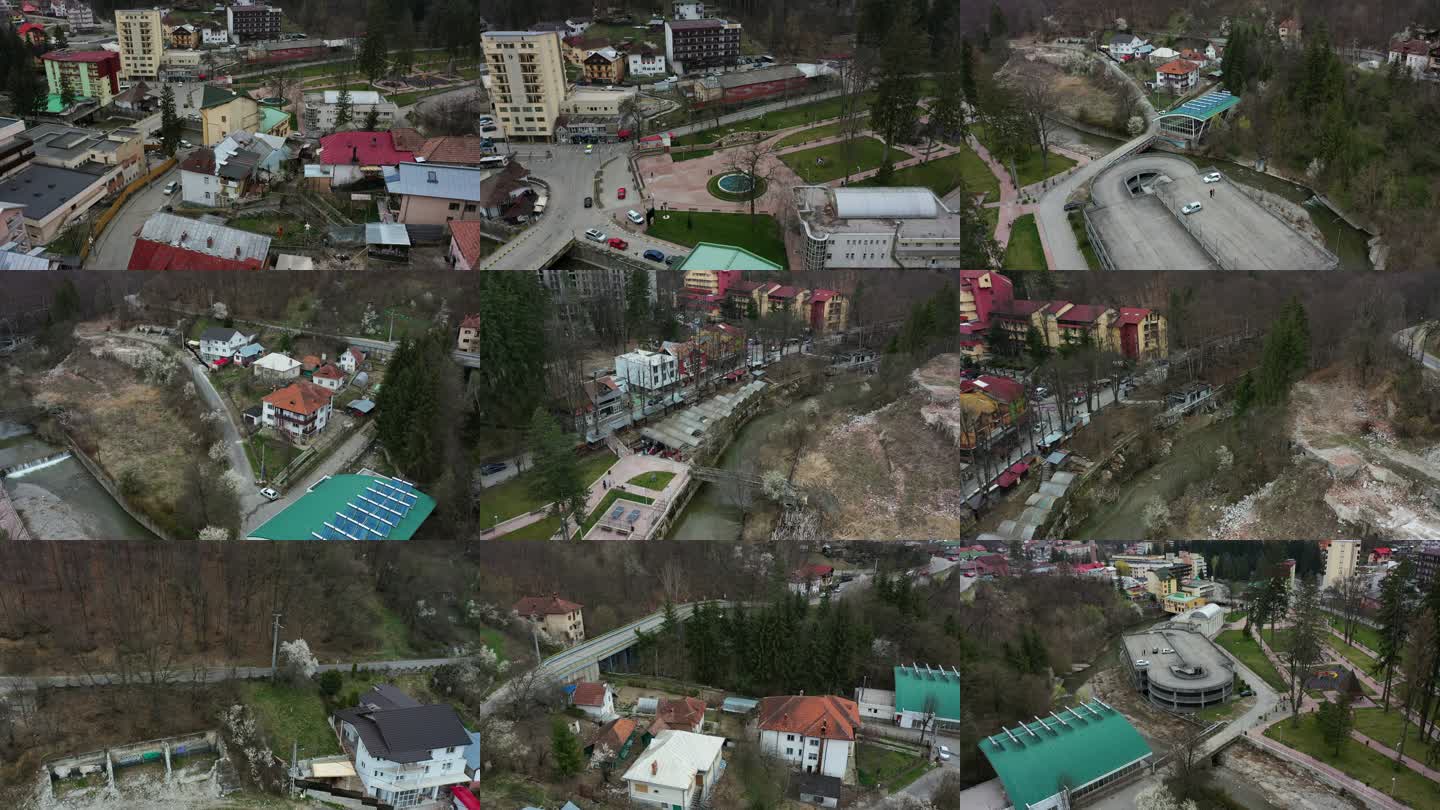}\\[2pt]
  \includegraphics[width=0.86\linewidth]{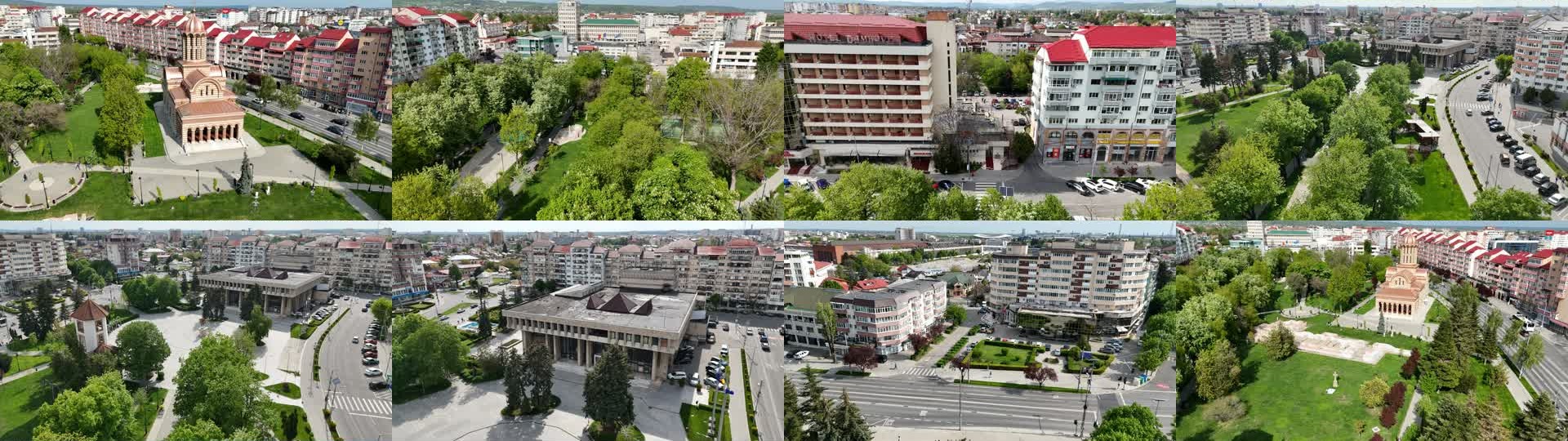}
  \caption{Sample drone viewpoints from the \textsc{olanesti} (top, held-out) and \textsc{targoviste} (bottom) scenes.}
  \label{fig:supp:sheets}
\end{figure}

\begin{figure}[t]
  \centering
  \includegraphics[width=0.94\linewidth]{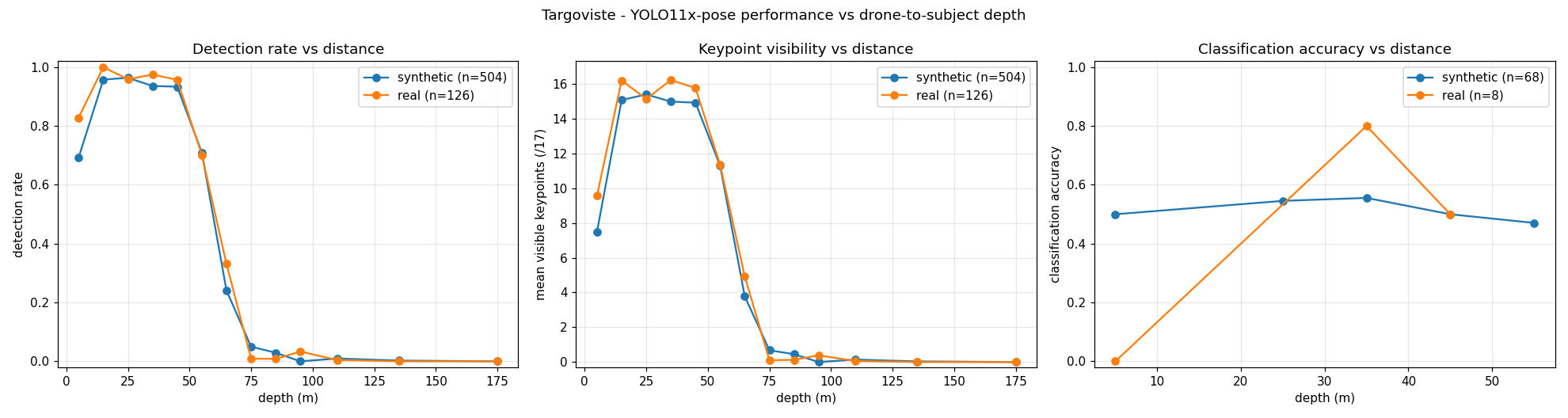}
  \caption{Single-scene (\textsc{targoviste}) pilot: YOLO11x-pose detection rate and mean visible keypoints vs.\ drone-to-subject depth, and classification accuracy of detected subjects. Detection, not classification, is the distance bottleneck.}
  \label{fig:supp:pilot}
\end{figure}

\Cref{fig:supp:sheets} shows sample drone viewpoints from two corpus scenes, illustrating the altitude, obliqueness and scene-structure variety the placement pipeline must handle. \Cref{fig:supp:pilot} reports the single-scene pilot study that motivated the distance band of the corpus: at full 4K, detection rate and keypoint visibility collapse beyond ${\sim}75$\,m, while classification accuracy on the detected subjects is already flat in distance -- the pattern the multi-scene study of the main text confirms.

\end{document}